\documentclass[conference]{IEEEtran}
\IEEEoverridecommandlockouts
\usepackage{cite}
\usepackage{amsmath,amssymb,amsfonts}
\usepackage{algorithmic}
\usepackage{graphicx}
\usepackage{textcomp}
\usepackage{xcolor}

\usepackage{graphicx}
\usepackage{amsmath}
\usepackage{amsfonts}
\usepackage{amssymb}
\usepackage[amssymb]{SIunits}
\usepackage{longtable}

\usepackage{import}
\usepackage{color}
\usepackage{amsthm}
\usepackage{cite}
\usepackage{subcaption}
\usepackage{textcomp}

\usepackage{lipsum}
\usepackage{multicol}
\usepackage{multirow}

\usepackage{scalerel}

\usepackage{algorithmic}
\usepackage{algorithm}
\usepackage{rotating}

\def\BibTeX{{\rm B\kern-.05em{\sc i\kern-.025em b}\kern-.08em
		T\kern-.1667em\lower.7ex\hbox{E}\kern-.125emX}}
\begin{document}
	
	\title{A Low Complexity Decentralized Neural Net with Centralized Equivalence using Layer-wise Learning\\
	}
	
	
	\author{\IEEEauthorblockN{Xinyue Liang, Alireza M. Javid, Mikael Skoglund, Saikat Chatterjee}
		\IEEEauthorblockA{\textit{School of Electrical Engineering and Computer Science} \\
			\textit{KTH Royal Institute of Technology}\\
			Stockholm, Sweden \\
			\{xinyuel, almj, skoglund, sach\}@kth.se}
	}
	
	\maketitle
	
	\begin{abstract}
		We design a low complexity decentralized learning algorithm to train a recently proposed large neural network in distributed processing nodes (workers). We assume the communication network between the workers is synchronized and can be modeled as a doubly-stochastic mixing matrix without having any master node. In our setup, the training data is distributed among the workers but is not shared in the training process due to privacy and security concerns. Using alternating-direction-method-of-multipliers (ADMM) along with a layer-wise convex optimization approach, we propose a decentralized learning algorithm which enjoys low computational complexity and communication cost among the workers. We show that it is possible to achieve equivalent learning performance as if the data is available in a single place. Finally, we experimentally illustrate the time complexity and convergence behavior of the algorithm.
	\end{abstract}
	

	\section{Introduction}

	Decentralized machine learning receives a high interest in signal processing, machine learning, and data analysis. In a decentralized setup, the training dataset is not in one place but distributed among several workers (or processing nodes). Due to physical limitations, the workers are connected with a communication network which is often represented as a graph in machine learning and signal processing fields. In such a communication network, data privacy and security among the workers are the main concerns in developing a decentralized learning algorithm. To this end, the following three aspects are of particular interest for a decentralized machine learning setup:
	\begin{enumerate}
		\item Workers are not allowed to share data, and there exists no master node that has access to all workers.
		\item The objective is to achieve the same performance as that of a centralized setup. 
		\item The learning algorithm should have a low computational complexity and communication overhead to efficiently handle large scale data.
	\end{enumerate}
	
	In this article, we develop a decentralized neural network for a classification problem to address these three aspects. The decentralized neural network is based on a recently proposed neural network called self-size estimating feedforward neural network (SSFN) \cite{SSFN_Saikat}. The SSFN is a multi-layer feedforward neural network that can estimate its size; meaning that the network automatically finds the necessary number of neurons and layers to achieve a certain performance. SSFN uses a rectified-linear-unit (ReLU) activation function and a special structure on the weight matrices. The weight matrices have two parts: one part is learned during the optimization process and the other part is predetermined as a random matrix instance. Weight matrices are learned using a series of convex optimization problems in a layer-wise fashion. The combination of layer-wise learning and the use of random matrices enables SSFN to be trained with a low computational requirement. Besides, the layer-wise nature of the training process leads to a significant reduction of communication overhead in decentralized learning compared to the gradient-based methods. Note that the SSFN does not use gradient-based methods, such as backpropagation, and hence does not require high computational resources. It is shown in \cite{SSFN_Saikat} that further optimization of weight matrices in SSFN using backpropagation does not lead to significant performance improvement.
	
	Our contribution is to develop a decentralized neural network by using the architecture and learning approach of SSFN that provides low computation and communication costs. We refer to this as decentralized SSFN (dSSFN) throughout the article. We use alternating-direction-method-of-multipliers (ADMM) \cite{Boyd_ADMM_2011} for finding decentralized solution of layer-wise convex optimization in dSSFN. Note that similar to \cite{SSFN_Saikat}, a decentralized estimation of the size of SSFN is possible in our framework as well, at the expense of higher complexity. In this article, we focus on training a fixed-size SSFN over a synchronous communication network. To seek consensus among the workers, we assume the communication network can be modeled by a doubly-stochastic mixing matrix. We conduct experiments for circular network topology, while our approach remains valid for sparse and connected communication networks as well. By systematically increasing the network connections between the workers, we investigate the trade-off between training time and the number of network connections. Besides, we experimentally show the convergence behavior of dSSFN throughout the layers and compare its classification performance against centralized SSFN for several well-known datasets.

	\subsection{Literature Review}
	
	There exists an enormous literature on distributed learning for large-scale data in recent years using huge computational resources \cite{DistBelief_2012, GUPTA2018, Jin2016, Anil2018}. The most prominent work in this area is the DistBelief framework which employs model parallelism techniques to use thousands of computing clusters to train a large neural network \cite{DistBelief_2012}. However, there is a growing need to develop algorithms that require less computational and communication resources. The use cases of such algorithms are internet-of-things, vehicular communication, sensor network, etc \cite{Li2018, Jiang2017}. 
	
	One popular approach to develop cost-efficient algorithms is to use variants of gradient-descent for distributed training of large neural networks. Stochastic gradient descent (SGD) and its variants, e.g., stochastic variance reduced gradient (SVRG), is designed to reduce the computational complexity of each iteration compared to the vanilla gradient descent \cite{SVRG2013}. Although these schemes are computationally efficient, they may significantly increase the communication complexity of the training process \cite{Bottou2018}. In particular, these approaches require a much larger number of iterations to ensure convergence to the true solution, and therefore, the number of information exchanges between the master node and each worker is potentially high. 
	
	This challenge has attracted wide attention in recent years. The approaches that are trying to address this issue can be seen as two different classes of algorithms. In the first class, a lossy quantization of the parameters and their gradient is employed to mitigate the huge communication burden, at the cost of a more number of iterations compared to the unquantized scheme. Some recent studies show that by carefully designing the quantizer at every step, it is possible to maintain the convergence speed of vanilla gradient descent \cite{Magnússon2019, Stich2018}. The second class of algorithms removes the requirement for master nodes to communicate with all workers at some iterations. In this way, the communication burden can be reduced at the cost of an increased local computational complexity \cite{Chen2018}. All of the above works investigate developing a cost-efficient algorithm in a master-slave topology and requires the communication to be synchronized. 
	
	Another widely studied algorithm for distributed optimization is the alternating direction method of multipliers (ADMM) and its variants. This class of algorithms has been studied by augmented Lagrangian methods or by operator theoretical frameworks \cite{Boyd_ADMM_2011, CoCoA2018, AROCK2016, Bastianello2019}. This class of algorithms gives more flexibility regarding the underlying topology and the required assumption on the communication links, e.g., synchronously and lossless communication. For example, \cite{AROCK2016} provides a framework for asynchronous updates of multiple workers under the assumption of having reliable communication links. \cite{Bastianello2019} extends this result and proposes a relaxed ADMM algorithm for asynchronous updates over lossy peer-to-peer networks and provides linear convergence near a neighborhood of the true solution. While the only gradient-based method that can deal with packet loss and partially-asynchronous updates is \cite{Alaviani2019} which implicitly requires the workers to use synchronized step-size \cite{Bastianello2019}. Thus, we choose ADMM as a different optimization approach to develop a cost-efficient distributed learning algorithm that gives us more flexibility regarding the underlying topology.
	
	There are several works for training artificial neural networks based on non-gradient algorithms \cite{Taylor2016, HUANG_ELM_2006, PLN_Saikat, DPLN_Xinyue}. \cite{Taylor2016} provides an ADMM-based method for joint training of all layers of a neural network. A fast yet effective architecture is random vector functional link (RVFL) networks that uses some of its parameters as randomly chosen between the input layer and the hidden layer while keeping direct links from the input layer to the output layer \cite{PAO_RVFL}. From the evaluation and proposed works of RVFL networks, it is observed that the non-iterative nature of RVFL leads to faster learning algorithms and low computational complexity in the distributed scenario \cite{ZHANG20161094, SCARDAPANE2015271}. A variant of RVFL is extreme learning machine (ELM) that removes the direct link between the input layer and the output layer while provides competitive performance with low complexity in different applications \cite{ELM_apps_1, ELM_apps_2, ELM_apps_3}.  in the distributed scenario. There have been several efforts to learn an ELM in a distributed manner. For example, He et.\ al. \cite{HE_MapReduce_ELM_2013} employs the advantages of Map-Reduce \cite{Dean_MapReduce_2008} to propose a distributed extreme learning machine scenario. We find a recent work \cite{LUO_DELM_NEUROCOMPUTING_2017} where they use ADMM to achieve the equivalent solution of the centralized ELM. In most of the works, they assume that every node in the network is fully connected to all other nodes.
	In this article, we investigate the network model in which every node has access to a limited number of neighbors.
	
	There exist works to develop deep randomized neural networks based on RVFL and its variants \cite{KATUWAL2019105854, Tang_HELM_2016, SSFN_Saikat}. They are shown to be capable of providing high-quality performance while keeping the computational complexity low. Katuwal et.\ al. \cite{KATUWAL2019105854} uses stacked autoencoders to construct a multi-layer RVFL network to obtain favorable performance while keeping low computational complexity. Tang et.\ al. \cite{Tang_HELM_2016} proposes the hierarchical ELM (H-ELM) which contains a multi-layer forward encoding part followed by the original ELM-based regression. The recent work by Chatterjee et.\ al. \cite{SSFN_Saikat} introduces a multi-layer ELM-based architecture called self size-estimating feed-forward neural network (SSFN). SSFN can estimate its size and guarantees the training error of the network to be decreasing as the number of layers increases. This is achieved using the lossless flow property \cite{SSFN_Saikat} and solving a constrained least-squares problem using ADMM at each layer. 
	
	In this article, we investigate the prospect of SSFN in a decentralized scenario over synchronous communication networks. The layer-wise nature of SSFN and the use of random weights makes SSFN an appealing option for low complexity design in distributed and online learning frameworks. Besides, the use of ADMM allows us to implement a decentralized SSFN with centralized equivalence \cite{Boyd_ADMM_2011}, while paves the way for extending this result to asynchronous and lossy communication networks \cite{AROCK2016, Bastianello2019} in our future studies.

	\section{Decentralized SSFN}
	
	We begin this section with a decentralized problem formulation for a feedforward neural network. Then, we briefly explain the architecture and learning of (centralized) SSFN followed by decentralization in synchronous communication networks. Finally, we show a comparison with a decentralized gradient descent algorithm. 
	
	\subsection{Problem formulation}
	
	In a supervised learning problem, let $(\mathbf{x},\mathbf{t})$ be a pair-wise form of data vector $\mathbf{x}$ that we observe and target vector $\mathbf{t}$ that we wish to infer. Let $\mathbf{x} \in \mathbb{R}^P$ and $\mathbf{t} \in \mathbb{R}^Q$. The target vector $\mathbf{t}$ can be a categorical variable for a classification problem with $Q$-classes.
	Let us construct a feed-forward neural network with $L$ layers, and $n_l$ hidden neurons in the $l$'th layer. We denote the weight matrix for $l$'th layer by $\mathbf{W}_l \in \mathbb{R}^{n_{l} \times n_{l-1}}$. For an input vector $\mathbf{x}$, a feed-forward neural network produces a mapping $\mathbf{f}: \mathbb{R}^P \rightarrow \mathbb{R}^{n_L}$ from input data to the feature vector in its last layer. The feature vector depends on parameters as $\mathbf{y} \triangleq \mathbf{f} \left(\mathbf{x}, \{ \mathbf{W}_l \}_{l=1}^L \right)$. 
	Then we use a linear transformation to generate target prediction as $\tilde{\mathbf{t}} = \mathbf{Oy}$, where $\mathbf{O} \in \mathbb{R}^{n_L \times Q}$ is the output matrix. We assume that there exists no parameter to optimize activation functions as they are predefined and fixed. A feed-forward neural network has the following form
	\begin{eqnarray*}
		\tilde{\mathbf{t}} = \mathbf{O} \mathbf{g}(\mathbf{W}_L \, \mathbf{g}(\hdots \mathbf{g}(\mathbf{W}_2 \, \mathbf{g}(\mathbf{W}_1 \, \mathbf{x}))\hdots)) = \mathbf{Oy},
	\end{eqnarray*}
	where $\mathbf{g}(\cdot)$ denotes the non-linear tranform function that uses a scalar-wise activation function, for example ReLU. The feedforward neural network signal flow follows sequenstial use of linear transform (LT) and non-linear transform (NLT).
	
	Suppose that we have a $J$-sample training dataset $\mathcal{D} = \{ (\mathbf{x}^{(j)},\mathbf{t}^{(j)} ) \}_{j=1}^J$.  The training dataset $\mathcal{D}$ is distributed over $M$ nodes in a decentralized setup as $\mathcal{D} = \cup_{m=1}^M \mathcal{D}_m$, where $\mathcal{D}_m$ denotes the dataset for $m$'th node. We assume that $\mathcal{D}_m \cap \mathcal{D}_n = \emptyset$. The dataset $\mathcal{D}_m$ is comprised of $J_m$ samples such that $\sum_{m=1}^M J_m = J$. 
	
	The output of the feed-forward neural network for the $m$'th node has the form $\tilde{\mathbf{t}}_m^{(j)} = \mathbf{O}_m \, \mathbf{f} (\mathbf{x}^{(j)}, \{ \mathbf{W}_{l,m} \} )$. The training cost for the $m$'th node is defined as
	\begin{eqnarray}
	\begin{array}{l}
	\mathcal{C}(\mathbf{O}_m, \! \{ \mathbf{W}_{l,m} \}) \triangleq \mathcal{C}(m)  \\  
	=  \displaystyle\sum_{(\mathbf{x}^{(j)},\mathbf{t}^{(j)} ) \in \mathcal{D}_m} \!\! \| \mathbf{t}^{(j)} - \mathbf{O}_m \, \mathbf{f} (\mathbf{x}^{(j)}, \{ \mathbf{W}_{l,m} \} ) \|^2,
	\end{array}
	\end{eqnarray}
	where $\| \cdot \|$ denotes $\ell_2$-norm of a vector. The total cost for the training dataset $\mathcal{D}$ over all nodes is $\sum_{m=1}^M \mathcal{C}(\mathbf{O}_m, \! \{ \mathbf{W}_{l,m} \})$.
	The decentralized learning problem is
	\begin{eqnarray}
	\begin{array}{r}
	\underset{\{ \mathbf{O}_m, \{ \mathbf{W}_{l,m}\} \} }{\arg\min}  \sum_{m=1}^M \mathcal{C}_m(\mathbf{O}_m, \! \{ \mathbf{W}_{l,m} \}) =  \sum_{m=1}^M \mathcal{C}(m) \\ \mathrm{s.t.} \,\,
	\left\{
	\begin{array}{l}
	\mathbf{W}_{l,m} = \mathbf{W}_l, \\
	\mathbf{O}_m = \mathbf{O}, \\
	\| \mathbf{W}_l \|_F^2 \leq \nu, \\
	\| \mathbf{O} \|_F^2 \leq \epsilon,
	\end{array}
	\right.
	\end{array}
	\label{eq:Decentralized_Optimization_Problem}
	\end{eqnarray}
	where $\mathbf{W}_{l,m} = \mathbf{W}_l$ and $\mathbf{O}_m = \mathbf{O}$ ensure that we have the same parameters for the set of neural networks across all $M$ nodes.            
	The constraints $\| \mathbf{W}_l \|_F^2 \leq \nu$ and $\| \mathbf{O} \|_F^2 \leq \epsilon$ are for regularization of parameters to avoid overfitting to the training dataset. Note that the constraints $\mathbf{W}_{l,m} = \mathbf{W}_l$ and $\mathbf{O}_m = \mathbf{O}$ lead to the case that the decentralized problem \eqref{eq:Decentralized_Optimization_Problem} is exactly equivalent to the following centralized problem
	\begin{eqnarray}
	\begin{array}{r}
	\underset{ \mathbf{O}, \{ \mathbf{W}_{l}\}  }{\arg\min} \,\, \mathcal{C}=\displaystyle\sum_{(\mathbf{x}^{(j)},\mathbf{t}^{(j)} ) \in \mathcal{D}} \!\! \| \mathbf{t}^{(j)} - \mathbf{O} \, \mathbf{f} (\mathbf{x}^{(j)}, \{ \mathbf{W}_{l} \} ) \|^2 \\ \mathrm{s.t.} \,\,
	\left\{
	\begin{array}{l}
	\| \mathbf{W}_l \|_F^2 \leq \nu, \\
	\| \mathbf{O} \|_F^2 \leq \epsilon,
	\end{array}
	\right.
	\end{array}
	\label{eq:Centralized_Optimization_Problem}
	\end{eqnarray}
	if the problem \eqref{eq:Centralized_Optimization_Problem} has a unique solution. It is well known that the above optimization problem is non-convex with respect to its parameters, and a learning algorithm will generally provide a suboptimal solution as a local minima.

	\subsection{Centralized SSFN}
\begin{figure*}[t!]
		\centering
		\def\svgwidth{\linewidth}
		\includegraphics[width=1\textwidth]{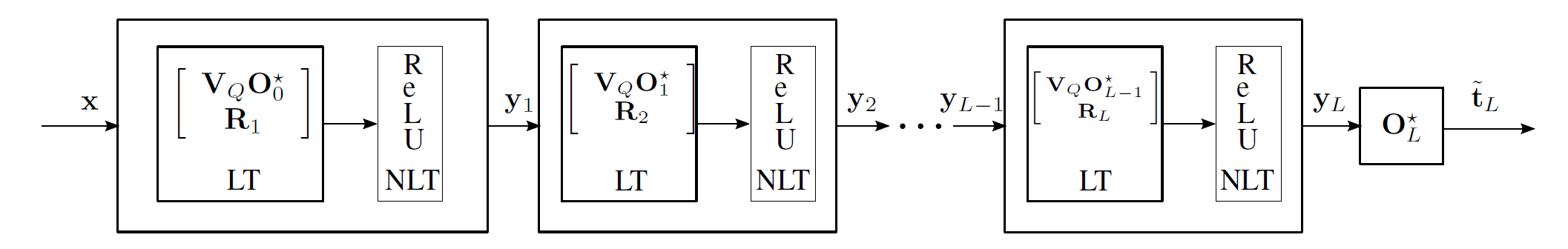}
		\caption{The architecture of a multi-layer SSFN with $L$ layers and its signal flow diagram. LT stands for \emph{linear transform} (weight matrix) and NLT stands for \emph{non-linear transform} (activation function). We use ReLU activation function.}
		\label{fig:MultiLayerPLN}
	\end{figure*}

	To design decentralized SSFN, we briefly discuss SSFN in this section for completeness. Details can be found in \cite{SSFN_Saikat}. SSFN is a feedforward neural network and its design requires a low computational complexity. The architecture of SSFN with its signal flow diagram is shown in Figure \ref{fig:MultiLayerPLN}. 
	\medmuskip=-2mu
	\medmuskip=4mu
	
	While the work of \cite{SSFN_Saikat} developed the SSFN architecture that learns its parameters and size of the network automatically, we work with a fixed size SSFN and learn its parameters. Note that our proposed method remains valid for estimating the size at the cost of higher complexity. The number of layers $L$ and the hidden neurons for the $l$'th layer $n_l$ are fixed a-priori. For simplicity, we assume that all layers have the same number of hidden neurons, which means $n_l =n, \forall l$.
	
	The SSFN addresses the optimization problem \eqref{eq:Centralized_Optimization_Problem} in a suboptimal manner. The SSFN parameters $\mathbf{O}$ and $\{\mathbf{W}_l\}$  are learned layer-by-layer in a sequential forward learning approach. The feature vector of $l$'th layer is constructed as follows 
	\begin{eqnarray}
	\mathbf{y}_l = \mathbf{g}(\mathbf{W}_l \, \mathbf{g}(\hdots \mathbf{g}(\mathbf{W}_2 \, \mathbf{g}(\mathbf{W}_1 \, \mathbf{x}))\hdots)) \in \mathbb{R}^n. 
	\end{eqnarray}
	The layer-by-layer sequential learning approach starts by optimizing layer number $l=1$ and then the new layers are added and optimized one-by-one until we reach $l=L$. Let us first assume that we have an $l$-layer network. The $(l+1)$-layer network will be built on an optimized $l$-layer network. We define $\mathbf{y}_0 = \mathbf{x}$. For designing the $(l+1)$-layer network given the $l$-layer network, the steps of finding parameter $\mathbf{W}_{l+1}$ are as follows:
	\begin{enumerate}
		\item For all the samples in the training dataset $\mathcal{D}$, we compute $\mathbf{y}_l^{(j)} = \mathbf{g}(\mathbf{W}_l \, \mathbf{g}(\hdots \mathbf{g}(\mathbf{W}_2 \, \mathbf{g}(\mathbf{W}_1 \, \mathbf{x}^{(j)}))\hdots))$.
		\item Using the samples $\{ \mathbf{y}_l^{(j)} \}_{j=1}^J$ we define a training cost 
		\begin{eqnarray}
		\mathcal{C}_l = \sum_{j=1}^J  \| \mathbf{t}^{(j)} - \mathbf{O}_l \, \mathbf{y}_l^{(j)} \|^2.
		\end{eqnarray} 
		We compute the optimal output matrix $\mathbf{O}_l$ by solving the convex optimization problem
		\begin{eqnarray}
		\begin{array}{r}
		\mathbf{O}_{l}^{\star} = \underset{ \mathbf{O}_{l}  }{\arg\min} \,\, \mathcal{C}_l  \,\,  \mathrm{s.t.} \,\,
		\| \mathbf{O}_l \|_F^2 \leq \epsilon_l.
		\end{array}
		\label{eq:LayerWise_ConvexOptimization}
		\end{eqnarray}
		It is shown in \cite{SSFN_Saikat} that we can choose the regularization parameters $\epsilon_l=\epsilon=2Q, \, l=0,1,2,\hdots,L.$ Note that $\mathbf{O}_0$ is a $Q \times P$-dimensional matrix, and every $\mathbf{O}_l$ for $l=1,2,\hdots,L$ is a $Q \times n$-dimensional amtrix. 
		\item We form the weight matrix for the $(l+1)$'th layer
		\begin{equation}
		\mathbf{W}_{l+1}=
		\left[
		\begin{array}{c}
		\mathbf{V}_Q \mathbf{O}_{l}^{\star} \\ \mathbf{R}_{l+1}
		\end{array}
		\right],
		\label{eq:weightMatrix}
		\end{equation}
		where $\mathbf{V}_Q=[\mathbf{I}_Q \,\, -\mathbf{I}_Q]^{T}$ is a fixed matrix of dimension $2Q \times Q$, $\mathbf{O}_{l}^{\star}$ is learned by convex optimization \eqref{eq:LayerWise_ConvexOptimization}, and 
		$\mathbf{R}_{l+1}$ is an instance of random matrix. The matrix $\mathbf{R}_0$ is $(n-2Q) \times P$-dimensional, and every $\mathbf{R}_l$ for $l=1,2,\hdots,L$ is $(n-2Q) \times n$-dimensional. Note that we only learn $\mathbf{O}_{l}^{\star}$ to form $\mathbf{W}_l$. We do not learn $\mathbf{R}_{l}$ as it is pre-fixed before training of SSFN. After constructing the weight matrix according to \eqref{eq:weightMatrix}, the $(l+1)$-layer network is
		\begin{eqnarray}
		\begin{array}{rl}
		\mathbf{y}_{l+1} & = \mathbf{g}(\mathbf{W}_{l+1} \, \mathbf{g}(\hdots \mathbf{g}(\mathbf{W}_2 \, \mathbf{g}(\mathbf{W}_1 \, \mathbf{x}))\hdots)) \\
		&= \mathbf{g}(\mathbf{W}_{l+1} \mathbf{y}_{l}).
		\end{array}
		\end{eqnarray}
	\end{enumerate}
	It is shown in \cite{SSFN_Saikat} that the three steps mentioned above guarantee monotonically decreasing cost $\sum_{j}  \| \mathbf{t}^{(j)} - \mathbf{O}_l \, \mathbf{y}_l^{(j)} \|^2$ with increasing the layer number $l$. The monotonically decreasing cost is the key to address the optimization problem \eqref{eq:Centralized_Optimization_Problem} as we continue to add new layers one-by-one and set the weight matrix of every layer using \eqref{eq:weightMatrix}. It was experimentally shown (see Table 5 of \cite{SSFN_Saikat}) that the use of gradient search (backpropagation) for further optimization of parameters in SSFN could not provide any noticeable performance improvement. Note that backpropagation-based optimization requires a significant computational complexity compared to the proposed layer-wised approach.
	
	\subsection{Decentralized SSFN for Synchronous Communication}
	\label{subsec:Decentralized_SSFN_using_ADMM_Synchronous}
	
	We now focus on developing decentralized SSFN (dSSFN) where information exchange between $M$ nodes follows synchronous communications. The main task is finding decentralized solution of the convex optimization problem~\eqref{eq:LayerWise_ConvexOptimization}. We recast the optimization problem~\eqref{eq:LayerWise_ConvexOptimization} in the following form
	\begin{eqnarray}
	\label{eq:decentralized_optimization_problem_sync}
	\begin{array}{r}
	\underset{\mathbf{O}_{l,m}, \mathbf{Z}}{\mathrm{min}} \displaystyle \sum_{m=1}^{M}  \sum_{(\mathbf{x}^{(j)},\mathbf{t}^{(j)})\in \mathcal{D}_m}  \| \mathbf{t}^{(j)} - \mathbf{O}_{l,m} \mathbf{y}_l^{(j)} \|^2 \,\, \\ 
	\mathrm{s.t.} \,\, \|\mathbf{Z}\|_F^2 \leq \epsilon , \forall m, \mathbf{O}_{l,m} = \mathbf{Z},
	\end{array}
	\end{eqnarray}
	where $\mathbf{Z}$ is an auxiliary variable. We use matrix notation henceforth for simplicity. For the $m$'th node on graph, we define the following matrices: $\mathbf{T}_m$ is a $Q \times J_m$-dimensional matrix comprising of the column vectors $\mathbf{t}^{(j)} \in \mathcal{D}_m$, $\mathbf{X}_{m}$ is a $P \times J_m$-dimensional matrix comprising of the column vectors $\mathbf{x}^{(j)} \in \mathcal{D}_m$, and $\mathbf{Y}_{l,m}$ is a $n \times J_m$-dimensional matrix comprising of the column vectors $\mathbf{y}_l^{(j)}$ in the $l$'th layer. The matrices $\mathbf{T}_m$, $\mathbf{X}_{m}$, and $\mathbf{Y}_{l,m}$ correspond to the dataset $\mathcal{D}_m$. Using the matrix notation, the optimization problem \eqref{eq:decentralized_optimization_problem_sync} can be written as
	\begin{eqnarray}
	\label{eq:ClsADMM}
	\begin{array}{r}
	\underset{\mathbf{O}_{l,m}, \mathbf{Z}}{\mathrm{min}} \sum_{m=1}^{M}  \| \mathbf{T}_{m} - \mathbf{O}_{l,m}\mathbf{Y}_{l,m} \|_F^2, \,\, \mathrm{s.t.} \,\, \|\mathbf{Z}\|_F^2 \leq \epsilon , \\ \forall m, \mathbf{O}_{l,m} = \mathbf{Z},
	\end{array}
	\end{eqnarray}
	where $\mathbf{Z}$ is an auxiliary variable. By using alternating direction method of multipliers (ADMM) \cite{Boyd_ADMM_2011}, we break it into three subproblems as follows
	\begin{eqnarray}
	\begin{array}{lll} 
	\mathbf{O}_{l,m}^{\star}=\underset{\mathbf{O}}{\mathrm{argmin}} \|\mathbf{T}_{m}\textendash\mathbf{O}\mathbf{Y}_{l,m}\|_F^2+\frac{1}{\mu_l}\|\mathbf{O}\textendash\mathbf{Z}+\mathbf{\Lambda}_{m}\|_F^2, \\
	\mathbf{Z}^{\star}=\underset{\mathbf{Z}}{\mathrm{argmin}}  \sum_{m=1}^{M} \|\mathbf{O}_{l,m}^{\star}\textendash\mathbf{Z}+\mathbf{\Lambda}_{m}\|_F^2 \,\, \mathrm{s.t.} \,\, \|\mathbf{Z}\|_F^2 \leq \epsilon,\\
	\mathbf{\Lambda}_{m}=\mathbf{\Lambda}_{m}+\mathbf{O}_{l,m}^{\star}-\mathbf{Z}^{\star}.
	\end{array}\nonumber
	\end{eqnarray}
	Here, $\mu_l$ is the Lagrangian parameter of ADMM in the $l$'th layer, and $\mathbf{\Lambda}_m$ is the scaled dual variable at node $m$. The ADMM iterations are:
	\begin{eqnarray}
	\begin{array}{rl}
	\mathbf{O}_{l,m}^{(k+1)}&=\big(\mathbf{T}_{m}\mathbf{Y}_{l,m}^T+\frac{1}{\mu_l}(\mathbf{Z}^{(k)}-\mathbf{\Lambda}_m^{(k)})\big) \\
	& \hspace{1.5cm} \times (\mathbf{Y}_{l,m}\mathbf{Y}_{l,m}^T+\frac{1}{\mu_l}\mathbf{I})^{-1}, \\
	\mathbf{Z}^{{(k+1)}}&=\mathcal{P}_{\epsilon}(\frac{1}{M}\sum_{m=1}^{M}(\mathbf{O}_{l,m}^{{(k+1)}}+\mathbf{\Lambda}_{m}^{(k)})), \\
	\mathbf{\Lambda}_{m}^{{(k+1)}}&=\mathbf{\Lambda}_{m}^{{(k)}}+\mathbf{O}_{l,m}^{{(k+1)}}-\mathbf{Z}^{{(k+1)}},
	\end{array}
	\label{eq:ADMM_iterations}
	\end{eqnarray}
	where $k$ denotes the iteration for ADMM, and $\mathcal{P}_{\epsilon}$ performs projection onto the space of matrices with Frobenius norm less than or equal to $\epsilon$. The operation $\mathcal{P}_{\epsilon}$ is defined as
	\begin{equation}
	\mathcal{P}_{\epsilon}(\mathbf{Z})=\left\{
	\begin{array}{ll}
	\mathbf{Z}\cdot(\frac{\epsilon}{\|\mathbf{Z}\|_F}) & : \|\mathbf{Z}\|_F> \epsilon\\
	\mathbf{Z} & : \mbox{otherwise}.
	\end{array} 
	\right.\nonumber
	\end{equation}
	
	For the $k$'th iteration of ADMM, it is required that the average quantity $\frac{1}{M}\sum_{m=1}^{M}(\mathbf{O}_{l,m}^{{(k+1)}}+\mathbf{\Lambda}_{m}^{(k)})$ be available to every node. This average can be found by seeking consensus over the graph.
	It can be easily seen that if the graph topology is modeled as a doubly-stochastic matrix, it is possible to achieve the consensus across all nodes by a sufficiently large number of exchanges throughout the network \cite{Boyd_Gossip_2005}. The main steps of decentralized SSFN are shown in Algorithm~\ref{alg:PLN_algorithm}.
	
	\begin{algorithm}[ht!]
		\caption{: Algorithm for learning decentralized SSFN}\label{alg:PLN_algorithm}
		\mbox{Input: }
		\begin{algorithmic}[1]
			\STATE Training dataset $\mathcal{D}_m$ for the $m$'th node
			\STATE Parameters to set: $L, \mu_0, \mu_l, n$
			\STATE Set of random matrices $\{ \mathbf{R}_l \}_{l=1}^L$ are generated and shared between all nodes 
		\end{algorithmic} 
		\mbox{Initialization:}
		\begin{algorithmic}[1]
			\STATE $l=-1$  \hfill (Index for $l$'th layer)
		\end{algorithmic}
		\mbox{Progressive growth of layers:}
		\begin{algorithmic}[1]
			\REPEAT 
			\STATE $l \leftarrow l+1$  \hfill (Increase layer number)
			\STATE $k=0$ \hfill (Iteration index of ADMM)
			\STATE Compute $\mathbf{Y}_{l,m} = \mathbf{g}(\mathbf{W}_l \hdots \mathbf{g}(\mathbf{W}_1\mathbf{X}_{l,m})\hdots) = \mathbf{g}(\mathbf{W}_l \mathbf{Y}_{l-1,m})$ \\
			\hspace{-22pt}\mbox{Solve \eqref{eq:LayerWise_ConvexOptimization} in decentralized form \eqref{eq:ClsADMM} to find $\mathbf{O}_l^{\star}$:}
			\REPEAT
			\STATE $k \leftarrow k+1$
			\STATE Solve $\mathbf{O}_{l,m}^{(k+1)}$ using \eqref{eq:ADMM_iterations}
			\STATE Find $\frac{1}{M}\sum_{m=1}^{M}(\mathbf{O}_{l,m}^{{(k+1)}}+\mathbf{\Lambda}_{m}^{(k)})$ using consensus over graph
			\STATE Find $\mathbf{Z}^{(k+1)}$ and $\mathbf{\Lambda}_{m}^{(k+1)}$ by \eqref{eq:ADMM_iterations}
			\UNTIL $k=K$
			\STATE Form weight matrix $\mathbf{W}_{l+1}=
			\left[
			\begin{array}{c}
			\mathbf{V}_Q \mathbf{O}_{l}^{\star} \\ \mathbf{R}_{l+1}
			\end{array}
			\right]$
			\UNTIL  $l = L$
		\end{algorithmic}
	\end{algorithm}

	\subsection{Synchronous communication}
	
	To guarantee that every node learns the same SSFN structure with centralized equivalence, it is required to have synchronous communication and computation over the graph. This synchronized manner is also used for exchanging $(\mathbf{O}_{l,m}+\mathbf{\Lambda}_{m})$ in $\mathbf{Z}$-update in equation \eqref{eq:ADMM_iterations}. After ADMM converges for all the nodes on the graph, we construct one more layer of SSFN and repeat until we learn the parameters for all the $L$ layers.

	\subsection{Comparison with decentralized gradient search}
	
	We now present a comparison with distributed gradient descent for neural networks. While being generally a powerful method, gradient descent has practical limitations due to a high computational complexity and communication overhead.
	Let us assume for simplicity that there is no regularization constraints on $\mathbf{W}_l$ and $\mathbf{O}$. Considering the weight matrix $\mathbf{W}_l$ at $l$'th layer of the neural network. The centralized gradient descent is 
	\begin{eqnarray}
	\mathbf{W}_l^{(i+1)} = \mathbf{W}_l^{(i)} - \kappa \frac{\partial \mathcal{C}}{\partial \mathbf{W}_l^{(i)}},
	\end{eqnarray}
	where $i$ denotes the iteration for gradient search and $\kappa$ is the step size of the algorithm.
	The centralized gradient descent can be done in the following decentralized manner:
	\begin{eqnarray}
	\begin{array}{rl}
	\mathbf{W}_l^{(i+1)} & = \frac{1}{M} \sum_{m=1}^M \mathbf{W}_{l,m}^{(i+1)} \\
	& = \frac{1}{M} \sum_{m=1}^M \left( \mathbf{W}_{l,m}^{(i)}  - \kappa \frac{\partial \mathcal{C}_m}{\partial \mathbf{W}_{l,m}^{(i)}} \right) \\
	& = \frac{1}{M} \sum_{m=1}^M \mathbf{W}_{l,m}^{(i)} - \kappa \frac{1}{M} \sum_{m=1}^M \frac{\partial \mathcal{C}_m}{\partial \mathbf{W}_{l,m}^{(i)}} \\
	& = \mathbf{W}_l^{(i)} - \kappa \frac{1}{M} \sum_{m=1}^M \frac{\partial \mathcal{C}_m}{\partial \mathbf{W}_{l,m}^{(i)}}. 
	\end{array}
	\end{eqnarray}
	For $i$'th iteration of gradient search, it is required that the average quantity $\frac{1}{M} \sum_{m=1}^M \frac{\partial \mathcal{C}_m}{\partial \mathbf{W}_{l,m}^{(i)}}$ be available to every node. An average can be found by seeking consensus over a communication graph. The communication property of such graphs can be modeled as a doubly-stochastic mixing matrix. Therefore, under the technical condition of consensus seeking, it is possible to realize decentralized gradient search which is exactly the same as the centralized setup. Assume that we require $B$ iterations of information exchange to calculate an average quantity. Then assuming that the gradient descent requires $I$ iterations to converge, we need $BI$ times of information exchange. In practice, $B$ is in order of hundreds and $I$ is in order of thousands. Since the $\mathbf{W}_l$ matrix contains $n_l n_{l-1}$ scalars, the total information exchange for learning $\mathbf{W}_l$ is 
	\begin{eqnarray}
	n_l n_{l-1} BI.
	\end{eqnarray}
	In practice, this total information exchange may be very large and lead to a high communication load. Further, as the sparsity level of the graph increases, the required number of information exchanges $B$ also increases, and that leads to a longer training time for gradient descent.
	
	With this limitation of gradient descent, we take a different approach. We use a structured neural network where parameters are learned using ADMM to solve a convex optimization problem. The use of ADMM allows fast and efficient optimization in the decentralized scenario. 
	
	We now quantify the communication load for decentralized SSFN. Let us assume that we require $B$ iterations of information exchange across the nodes to calculate an average quantity. Assuming that the ADMM requires $K$ iterations, we need $BK$ times of information exchange for learning $\mathbf{O}_l^{\star}$ and forming $\mathbf{W}_l$ according to equation (\ref{eq:weightMatrix}). The submatrix $\mathbf{R}_l$ in $\mathbf{W}_l$ is an instance of random matrix, and it is pre-defined across all nodes. In practice, $B$ and $K$ are both in the order of hundreds. The $\mathbf{O}_l^{\star}$ matrix has $Q n_{l-1}$ scalars. Hence, the total information exchange for learning $\mathbf{W}_l$ is 
	\begin{eqnarray}
	Q n_{l-1} BK.
	\end{eqnarray}
	The ratio of communication load between gradient descent and decentralized SSFN is
	\begin{eqnarray}
	\eta = \frac{n_l n_{l-1} BI}{Qn_{l-1} BK} = \frac{n_lI}{QK} \gg 1,
	\end{eqnarray}
	since in practice, we have $I \gg K$ and $n_l \gg Q$. 

	\section{Experimental Evaluation}
	\label{section:SimResult}
	In this section, we apply numerical experiments to evaluate the performance of the decentralized SSFN. We compare the performance of decentralized SSFN with centralized SSFN. We investigate how the training time differs versus the connectivity of the underlying network. We learn SSFN on a decentralized underlying network with the following topology and properties.

	\begin{figure}[t!]
		\centering
		\includegraphics[width=0.4\textwidth]{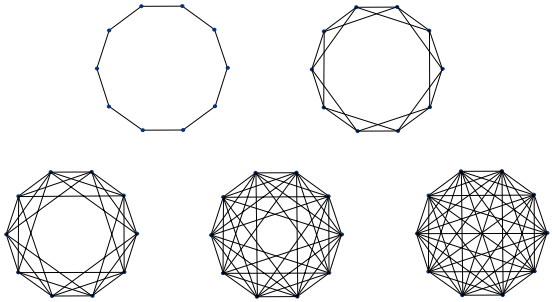}    
		\caption{Example of circular communication network topology. We have the number of nodes $M=10$ with degree $d=1, 2, 3, 4, 5$, respectively.}
		\vspace*{-10pt}
		\label{fig:NetworkDegree}
	\end{figure}
	
	\subsubsection{Network topology}
	
	The decentralized learning approach we propose in this manuscript can be performed on any network topology with the network mixing matrix being modeled as a doubly-stochastic matrix. There are several common types of network topology representations, such as $n$-connected cycles, random geometric structure, and $n$-regular expander structure \cite{DUCHI_DEFINITION_NETWORK_TOPOLOGY}. To perform a systematic study, we use circular topology as a communication network with a doubly-stochastic mixing matrix in the experiments. 
	
	Circular network topology with $M$ nodes has a degree $d$ to represent its connectivity. We show examples of circular topology in Figure \ref{fig:NetworkDegree}. A network with a degree $d$ has $d$ level of connected cycles between the neighbors. This implies that each node in the network has connections with $d$ neighbors on the left and right sides, respectively. A network with a low degree is considered sparse in the sense of having much fewer connections. A low degree in a network limits the number of information exchanges and subsequently affects the convergence speed of a decentralized learning algorithm. It is expected that a network consensus can be achieved faster if the degree of a graph increases.
	
	The communication property over the network can be modeled as a doubly-stochastic matrix $\mathbf{H}=[h_{ij}]_{M \times M}$ in which $h_{ij}$ is the weight of importance that the $i$'th node assigns to the $j$'th node during parameters exchange. The doubly-stochastic matrix has the following property:
	\begin{eqnarray*}
		\begin{array}{lll}
			h_{ij}=h_{ji} 
			\begin{cases}
				> 0, \hspace{20pt}  j \in \mathcal{N}_i\\
				= 0, \hspace{20pt}  j \not\in \mathcal{N}_i
			\end{cases}\! \!, \,\,\,
			\sum_{j=1}^{M}h_{ij}=\sum_{i=1}^{M}h_{ij}=1.
		\end{array}
	\end{eqnarray*}
	Here $\mathcal{N}_i$ refers to the the set of neighbors with whom the $i$'th node is connected. Note that $i \in \mathcal{N}_i$. In this setup, we assume that there is no master node and no node is isolated either. For the sake of simplicity, in the following experiments, the doubly-stochastic mixing matrix  is chosen in such a way that every node is connected to its neighbors with equal weights. That means we have $h_{ij} = \frac{1}{|\mathcal{N}_i|}$, where $|\mathcal{N}_i|$ denotes cardinality of the set $\mathcal{N}_i$. As we use circular network topology for the experiments, we have the relation 
	\begin{eqnarray*}
		\begin{array}{lll}
			|\mathcal{N}_i|= 
			\begin{cases}
				2d+1, \hspace{20pt}  d < d_{max}\\
				M, \hspace{20pt}  d = d_{max}
			\end{cases}\! \!, \,\,\,
		\end{array}
	\end{eqnarray*}
	for a graph with degree $d$.  
	
	\subsection{Classification tasks and datasets}
	We evaluate the decentralized SSFN for different classification tasks. 
	The datasets that we use are briefly mentioned in Table \ref{table:Database_for_classification}. 
	We use the $Q$-dimensional target vector $\mathbf{t}$ in a classification task represented as a one-hot vector. A target vector has only one scalar component that is 1, and the other scalar components are zero. 

	\begin{table}[t!]
		\centering
		\caption{Dataset for multi-class classification.}
		\vspace{-5pt}
		\label{table:Database_for_classification}
		\setlength{\tabcolsep}{2.5pt}
		\begin{tabular}{ |c|c|c|c|c| } 
			\hline
			Dataset & {\begin{tabular}{@{}c@{}}$\#$ of  \\ train data\end{tabular}}  & {\begin{tabular}{@{}c@{}}$\#$ of  \\ test data\end{tabular}} & {\begin{tabular}{@{}c@{}}Input  \\ dimension ($\mathit{P}$)\end{tabular}}  & {\begin{tabular}{@{}c@{}}$\#$ of  \\ classes ($\mathit{Q}$)\end{tabular}} \\
			\hline \hline 
			Vowel & 528 & 462 & 10 & 11 \\ 
			\hline
			Satimage & 4435 & 2000 & 36 & 6 \\ 
			\hline
			Caltech101 & 6000 & 3000 & 3000 & 102 \\ 
			\hline
			Letter & 13333 & 6667 & 16 & 26 \\ 
			\hline
			NORB & 24300 & 24300 & 2048 & 5 \\ 
			\hline
			MNIST & 60000 & 10000 & 784 & 10 \\ 
			\hline
		\end{tabular}
	\end{table}

	\subsection{Experimental setup}
	In all experiments, we set the number of layers $L=20$ and the number of hidden neurons $n=2Q+1000$ for each layer. We fix the number of nodes $M=20$ and uniformly divide the training dataset between the nodes. We set the number of iterations in ADMM as $K=100$ for each layer.
	
	\begin{table*}[t]
		\centering
		\caption{Classification performance comparison between centralized SSFN and decentralized SSFN on a circular communication network where $d=4$. 
		}
		\vspace{5pt}
		\label{table:Classification_accuracy}
		\setlength{\tabcolsep}{10pt}
		\begin{tabular}{|c|c|c|c|c|c|c|c|c|c|c|}
			\hline
			\multirow{2}{*}{Dataset} & \multicolumn{5}{c}{Centralized SSFN} & \multicolumn{5}{|c|}{Decentralized SSFN} \\ \cline{2-11}
			&{\begin{tabular}{@{}c@{}}Train \\ Accuracy\end{tabular}} & {\begin{tabular}{@{}c@{}}Train \\ Error\end{tabular}} & {\begin{tabular}{@{}c@{}}Test \\ Accuracy\end{tabular}} & {\begin{tabular}{@{}c@{}}$\mu_0$\end{tabular}}  &
			{\begin{tabular}{@{}c@{}}$\mu_l$\end{tabular}}  & 
			{\begin{tabular}{@{}c@{}}Train \\ Accuracy\end{tabular}} & {\begin{tabular}{@{}c@{}}Train \\ Error\end{tabular}} & {\begin{tabular}{@{}c@{}}Test \\ Accuracy\end{tabular}} & 
			{\begin{tabular}{@{}c@{}}$\mu_0$\end{tabular}}  &
			{\begin{tabular}{@{}c@{}}$\mu_l$\end{tabular}} \\
			\hline \hline
			
			Vowel & 100$\pm$0.00 & -53.8 & 58.3$\pm$1.70 & $10^{-3}$ & $1$ & 100$\pm$0.00 & -51.67 & 59.2$\pm$1.10 & $10^{-3}$ & $10^{1}$  \\ 
			\hline
			Satimage & 94.2$\pm$0.21 & -10.6 & 86.9$\pm$0.37 & $10^{-6}$ & $ 10^1 $ & 92.1$\pm$0.10 & -9.37 & 88.8$\pm$0.08 & $10^{-4}$ & $10^{-1}$ \\ 
			\hline
			Caltech101 & 99.9$\pm$0.01 & -38.9 & 73.2$\pm$0.91 & $10$ & $1$ & 99.9$\pm$0.01 & -34.94 & 75.4$\pm$0.29 & $10^{-1}$ & $10^{0}$ \\ 
			\hline
			Letter & 99.4$\pm$0.02 & -19.5 & 91.8$\pm$0.23 & $10^{-4}$ & $10$ & 98.9$\pm$0.03 & -17.64 & 92.5$\pm$0.22 & $10^{-6}$ & $10^{0}$  \\ 
			\hline
			NORB & 96.7$\pm$0.04 & -13.9 & 82.5$\pm$0.22 & $10^{-1}$ & $10^{-1}$ & 96.7$\pm$0.02 & -13.93 & 82.6$\pm$0.16 & $10^{-2}$ & $10^{0}$  \\ 
			\hline  
			MNIST & 96.8$\pm$0.06 & -12.9 & 94.8$\pm$0.16 & $10^{-4}$ & $10^{-1}$ & 97.0$\pm$0.04 & -13.24 &95.1$\pm$0.16 & $10^{-5}$ & $10^0$ \\ 
			\hline
		\end{tabular}
	\end{table*}
	
	\subsection{Experimental results}

	\begin{figure*}[th!]
		\centering
		\begin{multicols}{3}
			\includegraphics[width=0.35\textwidth]{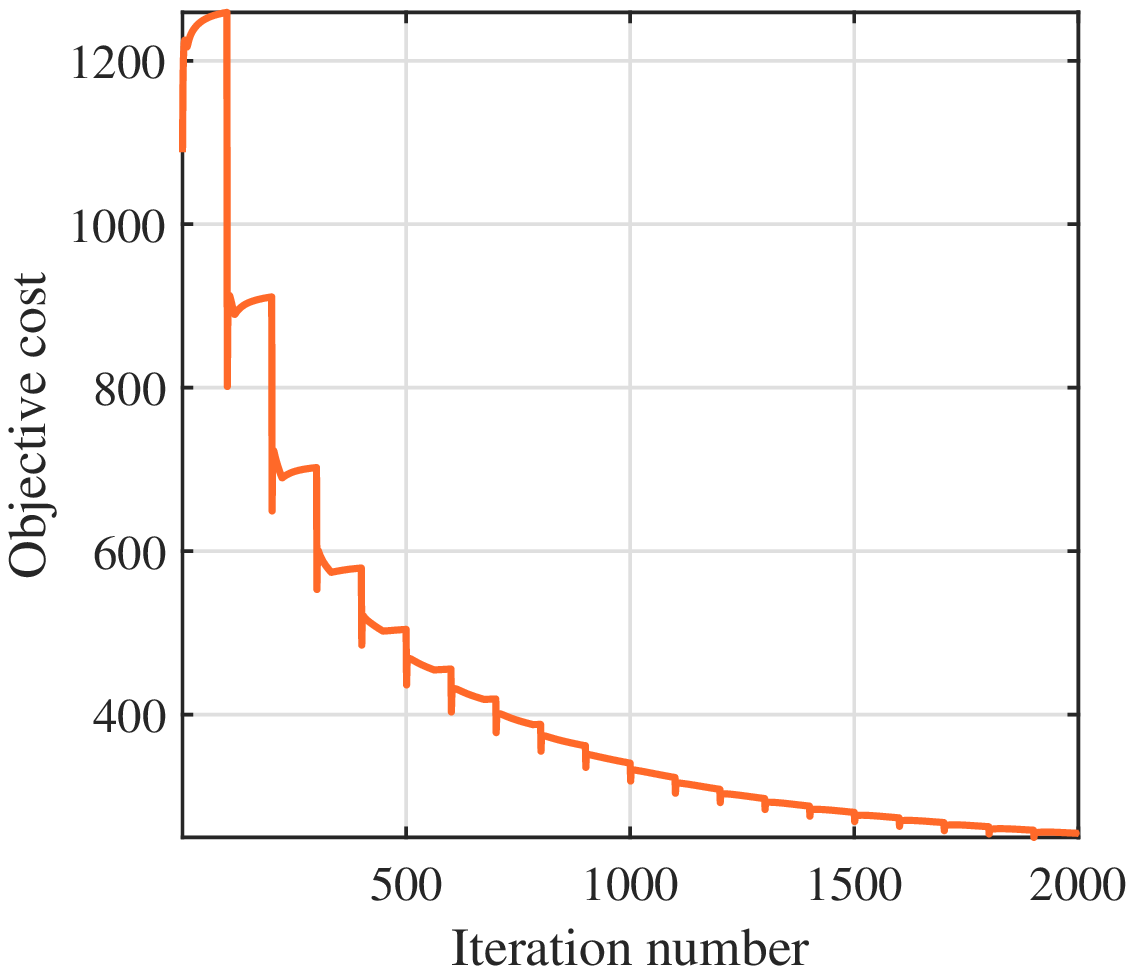}
			\subcaption{Satimage}
			\includegraphics[width=0.35\textwidth]{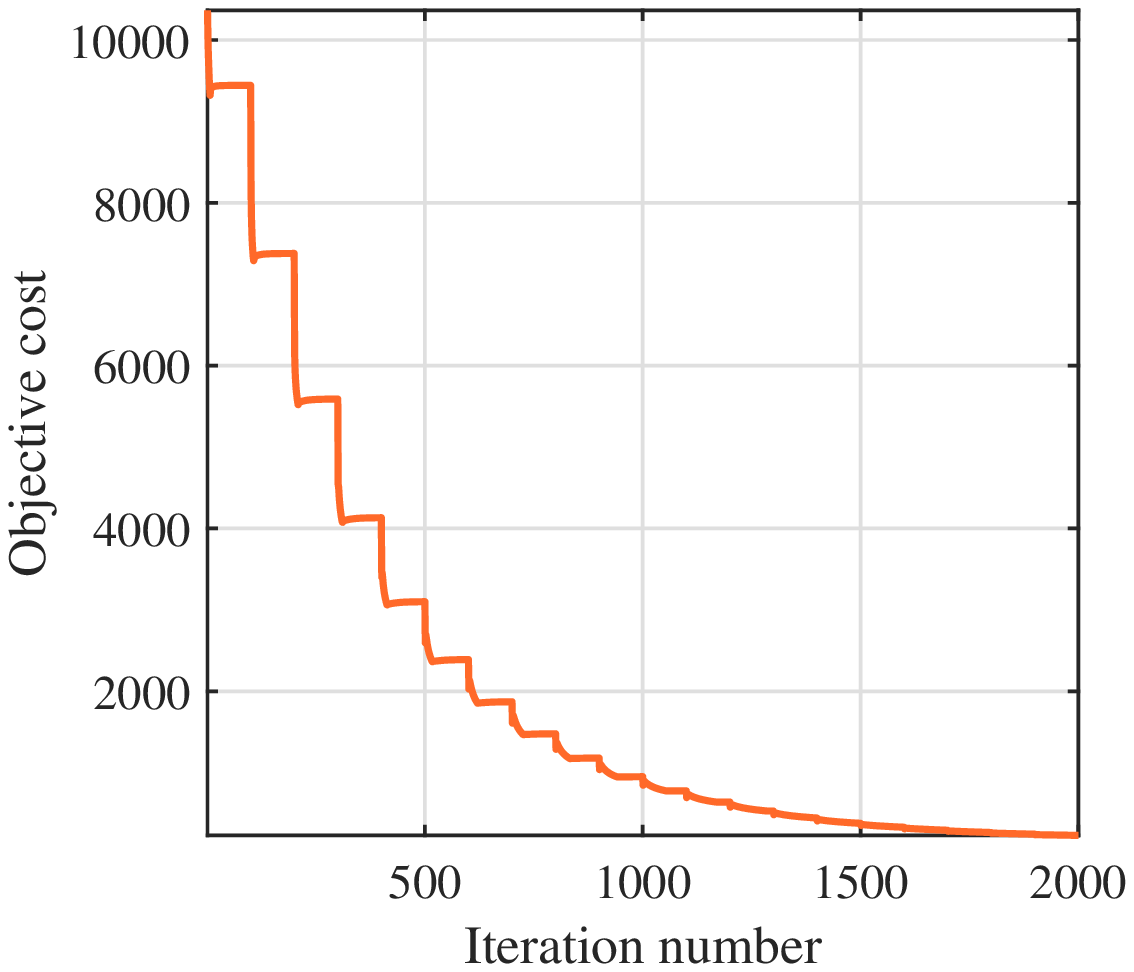}
			\subcaption{Letter}
			\includegraphics[width=0.35\textwidth]{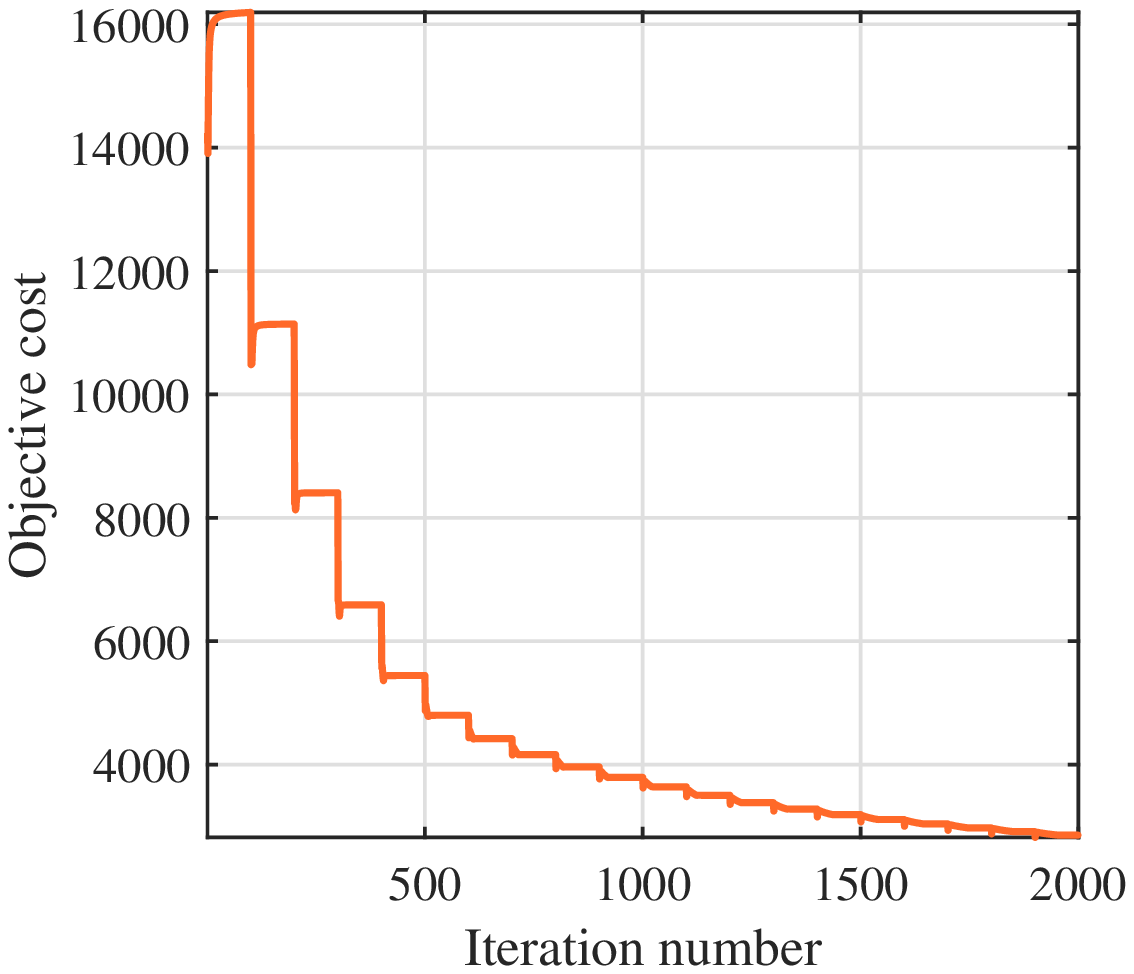}
			\subcaption{MNIST}
		\end{multicols}
		\vspace{-0.3cm}    
		\caption{Objective cost versus total number of ADMM iterations throughout all layers.}
		\label{fig:objective_cost}
		\begin{multicols}{3}
			\includegraphics[width=0.35\textwidth]{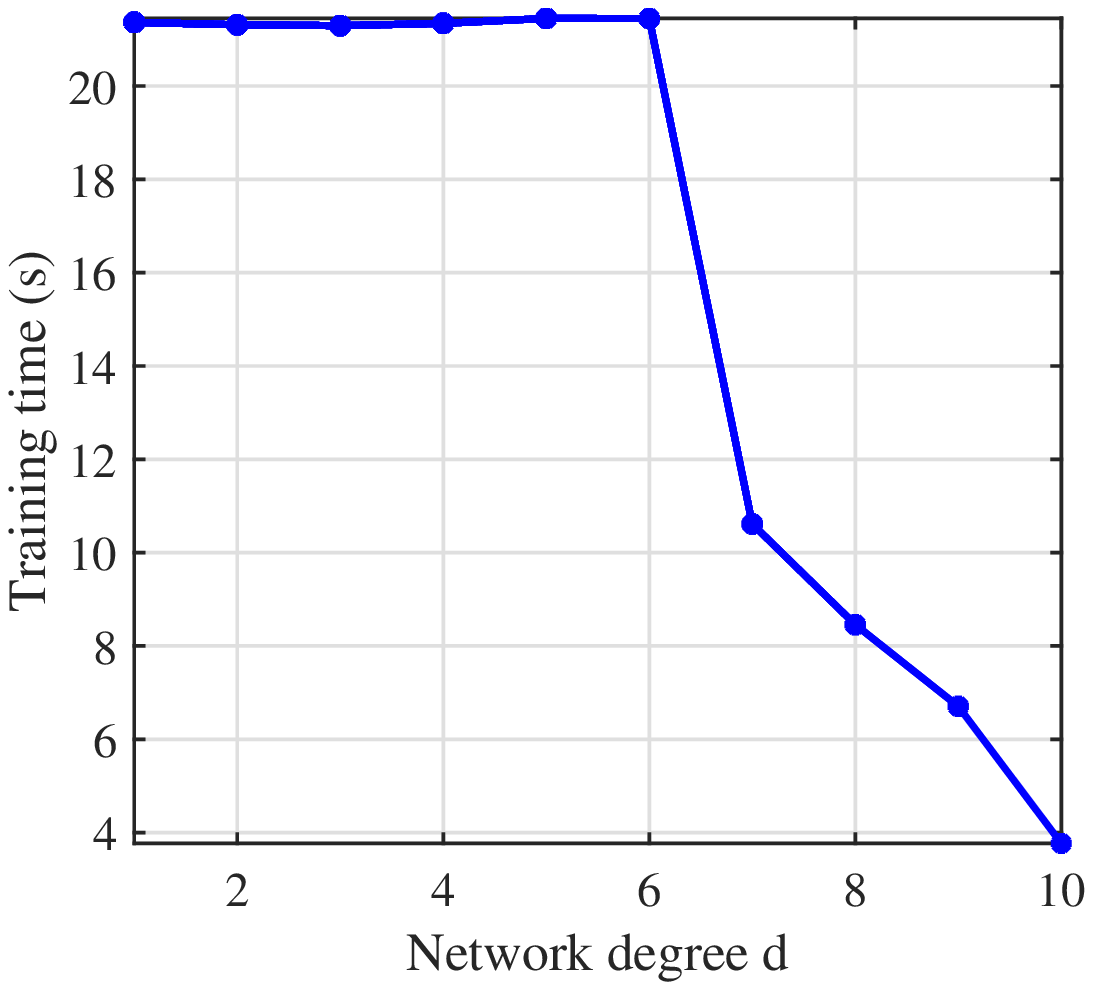}
			\subcaption{Satimage}
			\includegraphics[width=0.35\textwidth]{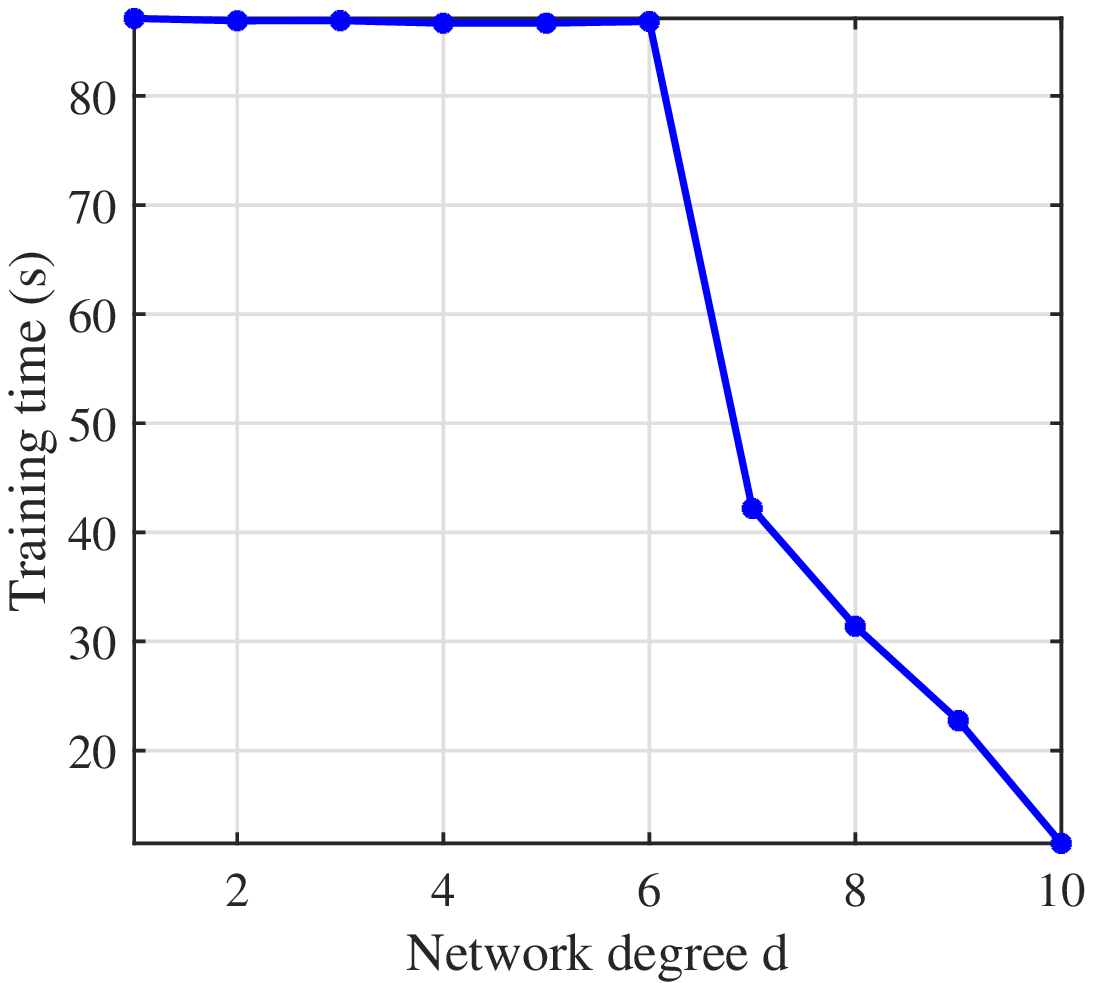}
			\subcaption{Letter}
			\includegraphics[width=0.35\textwidth]{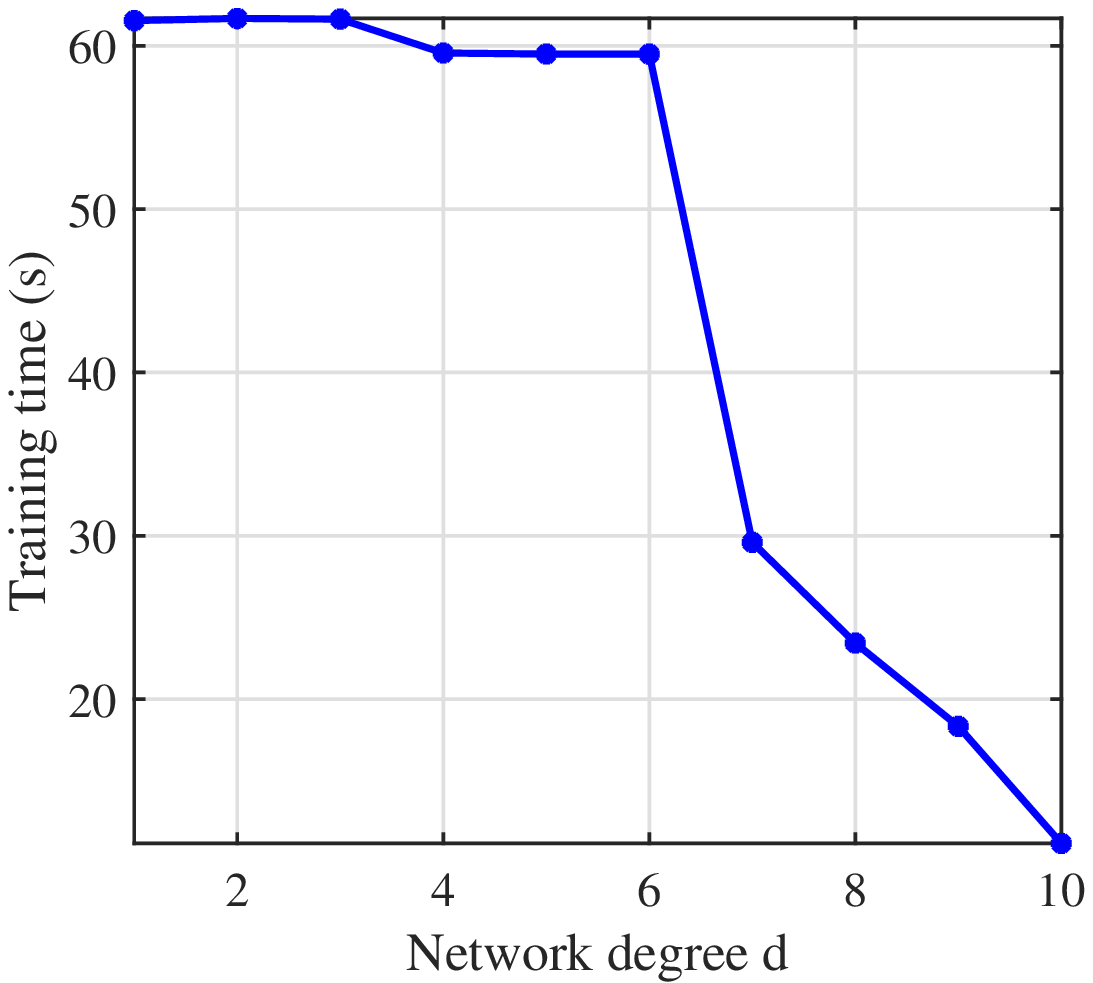}
			\subcaption{MNIST}
		\end{multicols}
		\vspace{-0.3cm}    
		\caption{Training time changes as the network degree increases on the $20$-node circular communcication network.}
		\label{fig:training_time}
	\end{figure*}

	We first show the performance of decentralized SSFN for a graph with a degree $d=4$ compared with the centralized SSFN. The performances are shown in Table \ref{table:Classification_accuracy}. It can be seen that dSSFN provides similar performance to centralized SSFN for the proper choice of hyperparameters. The practical performance of the decentralized SSFN is affected by the choice of hyperparameters $\mu_0, \mu_l$, the number of ADMM iterations $K$. Choosing proper $\mu_0$ and $\mu_l$ guarantees ADMM to converge within $K=100$ iterations. 
	
	The convergence behavior of dSSFN throughout the layers is shown in Figure \ref{fig:objective_cost}. The decentralized objective cost versus the total number of ADMM iterations in all layers is shown for Satimage, Letter, and MNIST dataset. For each layer (every $100$ ADMM iterations), ADMM converges to a global solution for the optimization problem \eqref{eq:ClsADMM}. Overall it can be observed that the curves show a power-law behavior. Similar to SSFN, the objective cost converges as we increase the number of layers. Therefore, we can decide to stop the addition of new layers when we see that the cost has a convergence trend.
	
	Figure \ref{fig:training_time} shows training time for learning decentralized SSFN versus network degree $d$ for Satimage, Letter, and MNIST datasets. It is interesting to observe that the training time shows a transition jump in the middle range of $d$. There exists a $d$ threshold after which the learning mechanism in decentralized SSFN converges noticeably faster. The degree represents sparsity in the graph, and in turn, relates to privacy, security, and physical communication links. Our results imply that a suitable network degree helps to achieve a trade-off between the graph degree and training time.

	\subsection{Reproducible codes}
	Matlab codes of all the experiments described in this paper are available at https://sites.google.com/site/saikatchatt/. The datasets used for the experiments can be found at \cite{DATABASE_1,JIANG_DICT_LEARNING_KSVD_CVPR_2011,Lecun_NORB_online,Lecun_MNIST_online} .

	\section{Conclusion}
	We develop a decentralized multilayer neural network and show that it is possible to achieve centralized equivalence under some technical assumptions. While being sub-optimal because of its layer-wise nature, the proposed method is cost-efficient compared to the general gradient-based methods in the sense of computation and communication complexities. We experimentally show the convergence behavior of dSSFN throughout the layers and provide a monotonically decreasing training cost by adding more layers. Besides, we inspect the time complexity of the algorithm under different network connectivity degrees. Our experiments show that dSSFN can provide centralized performance for a network with a high sparsity level in its connections. The proposed method is limited to the network topologies with a doubly-stochastic mixing matrix and synchronized connections. Extending this result to asynchronous and lossy peer-to-peer networks by using relaxed ADMM approaches is a potential future direction.

	\bibliographystyle{IEEEtran}
	\bibliography{biblio_DPLN}

\end{document}